# White Paper: Inverse Signal Classification for Financial Instruments

Uri Kartoun[1]

*2501 Porter St. #806, Washington D.C., 20008, U.S.A.*

*Stockato LLC*

+1-202-374-4007

uri@stockato.com

**Abstract.** The paper presents new machine learning methods: *signal composition*, which classifies time-series regardless of length, type, and quantity; and *self-labeling*, a supervised-learning enhancement. The paper describes further the implementation of the methods on a financial search engine system using a collection of 7,881 financial instruments traded during 2011 to identify inverse behavior among the time-series.

## 1. Introduction

Classification methods for financial instruments such as exchange-traded-funds (ETFs) and stocks, are commonly used to identify investments that meet one's personal criteria. Such methods aim to save time by narrowing one's search to a manageable number of specific investments for further research and examination. These classification methods (e.g., financial instrument screeners) facilitate a user to create

Kartoun U., Inverse Signal Classification for Financial Instruments, 2013.

a list of specific financial instruments he or she desires to further compare and analyze.

One disadvantage of current financial instrument classification systems is the lack of ability to classify different financial instruments based on similarities in behavior patterns. An example of a behavior pattern would be a time-series of a financial instrument considered in a specific time period, wherein the time-series is a sequence of data points that represent the daily change in the financial instrument price. Another disadvantage is the inability to classify financial instruments from different classes, for example, to find behavioral similarities between a certain stock and a certain ETF. Another disadvantage is the inability to classify financial instruments from different stock exchanges and/or from different countries, for example, to find behavioral similarities between a certain *NASDAQ*[2] stock and an *AMEX*[3] ETF. Another disadvantage is the inability to identify financial instrument with an inverse behavior. Identifying inverse behavior among financial instruments is especially useful to research alternative investments to diversify one's investment portfolio.

Varieties of methods have been used to classify time-series. Perng et al., 2000, propose the *Landmark Model* for similarity-based pattern querying in time-series databases - a model of similarity that is consistent with human intuition and episodic memory. *Landmark Similarity* measures are computed by tracking different specific subsets of features of landmarks. The authors report on experiments using 10-year closing prices of stocks in the *Standard & Poor 500* index. Nguyen et al., 2011, propose an algorithm called *LCLC (Learning from Common Local Clusters)* to create a classifier for time-series using limited labeled positive data and a cluster chaining approach to improve accuracy. The authors compare the *LCLC* algorithm with two existing semi-supervised methods for time-series classification: Wei's method (Wei and Keogh, 2006), and Ratana's method (Ratanamahatana and Wanichsan, 2008). To demonstrate the superiority of *LCLC*, the authors used five data-sets of time-series (Wei, 2007; Keogh, 2008).

Lines et al., 2012, propose a shapelet transform for time-series classification. Their implementation includes the development of a caching algorithm to store shapelets, and to apply a parameter-free cross-validation approach for extracting the most significant shapelets. Experiments included the transformation of 26 data-sets to demonstrate that a *C4.5* decision tree classifier trained with transformed data is competitive with an implementation of the original shapelet decision tree of Ye and Keogh, 2009. Lines et al., 2012, demonstrate that the filtered data can be applied also to non-tree based classifiers to achieve improved classification performance, while maintaining the interpretability of shapelets. Another signal classification approach is presented in Povinelli et al., 2004 - the approach is based upon modeling the dynamics of a system as they are captured in a reconstructed phase space. The modeling is based on full covariance Gaussian Mixture Models of time domain signatures. Three data-sets were used for validation, including motor current simulations, electro-cardiogram recordings, and speech waveforms. The approach is different than other signal classification approaches (such as linear systems analysis using frequency content and simple non-linear machine learning models such as artificial neural networks). The results demonstrate that the proposed method is robust across these diverse domains, outperforming the time delay neural network used as a baseline. Using artificial neural networks for classifying time-series, however, as described in (Haselsteiner and Pfurtscheller, 2000) proved to be robust - the authors address classification of electroencephalograph (EEG) signals using neural networks. The paper compares two topologies of neural networks. Standard multi-layer perceptrons (MLPs) are used as a method for classification, and are compared to finite impulse response (FIR) MLPs, which use FIR filters instead of static weights to allow temporal processing inside the classifier. Experiments with three different subjects demonstrate the higher performance of the FIR MLP compared with the standard MLP. Another example for using supervised learning (recurrent neural networks) for classifying time-series is provided as in (Hüsken and Stagge, 2003).

---

[2] National Association of Securities Dealers Automated Quotations.
[3] American Stock Exchange.

Jović *et al.*, 2012, examined the use of decision tree ensembles in biomedical time-series classification. Experiments performed focused on biomedical time-series data-sets related to cardiac disorders, demonstrated that the use of decision tree ensembles provide superior results in comparison with support vector machines (SVMs). In particular, *AdaBoost.M1* and *MultiBoost* algorithms applied to *C4.5* decision tree found as the most accurate.

The structure of the paper is as follows: after introducing the financial search engine system in Section 2, the time-series classification methods are detailed in Section 3. Conclusions are provided in Section 4.

## 2. A Financial Search Engine

### 2.1 Human-Computer Interaction

To receive a list of inversely behaving financial instruments, the user specifies a financial instrument at the user interface. The financial instrument is sent to query a classifying database and a list of financial instruments is received. The list contains one or more financial instruments that found to have inverse behavior patterns to the financial instrument specified. The list is sorted according to level of similarity criterion and presented at the user interface. As an example, a user may specify *PBR*. Immediately acquired from the database *HSY*, that was found with an inverse behavior to the specified financial instrument during 2011. Figure 1 depicts the user interface for the financial search engine.

Kartoun U., Inverse Signal Classification for Financial Instruments, 2013.

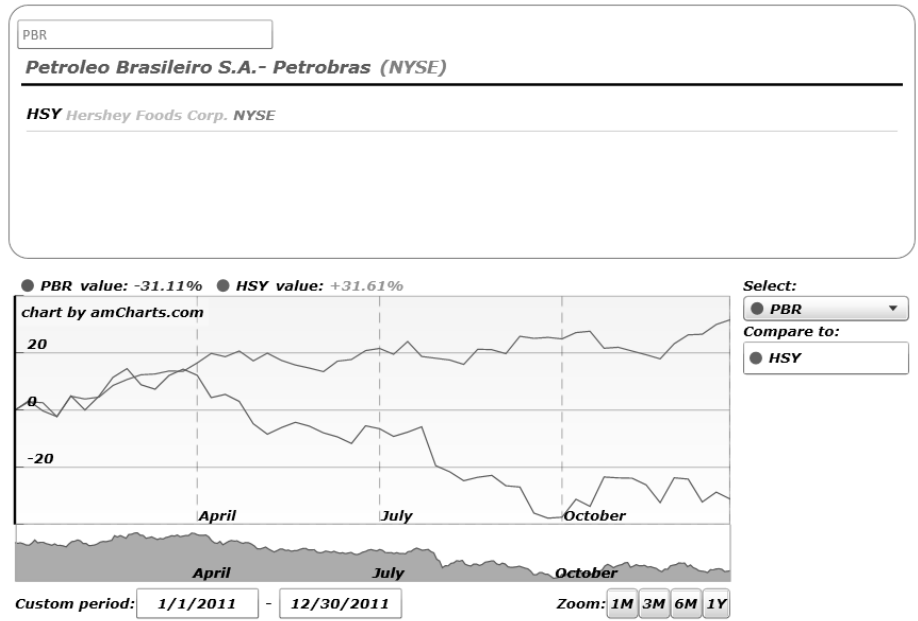

*Figure 1.* User interface.

## 2.2 Classification Module

To classify financial instruments, a classification procedure is used as shown in figure 2. The classification procedure is facilitated to perform a method that generates classification results. The available patterns are end-of-day prices of all financial instruments traded at *NASDAQ*, *NYSE*[4], and *AMEX* during 2011, including values of several market indices (*e.g.*, *DJI*, *FTSE*, *HIS*). The patterns are modified and sliced using a data preparation procedure as described through expressions 3.1 - 3.6. A decision tree learning algorithm is applied on the patterns for each time slice, and classification results are stored at a classifying database.

---

[4] New York Stock Exchange.

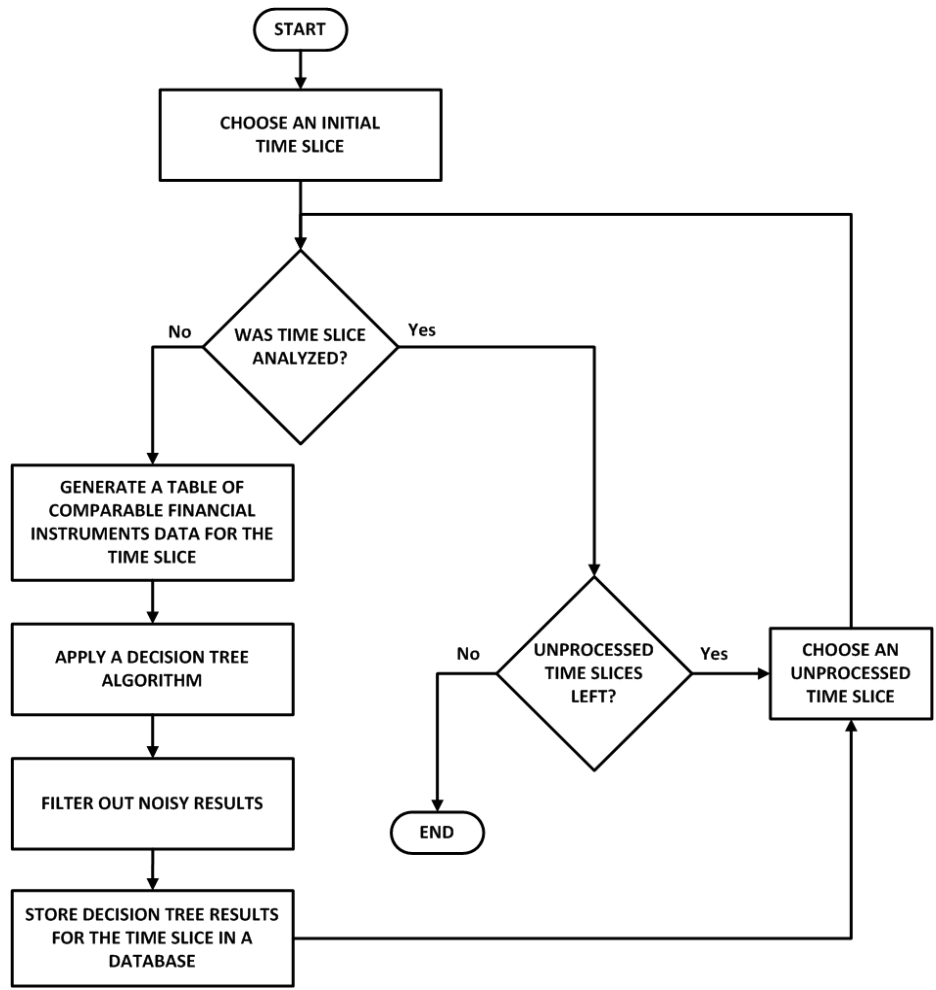

*Figure 2.* Classification procedure.

# 3. Time-series Classification

## 3.1 Self-Labeling

Assume $S_1, S_2, S_i, ... S_m$ are $m$ financial instruments considered for classification during a trading time-range that includes $n$ time-steps (*e.g.*, a one time-step equals to a one day). Each financial instrument $S_i$ is associated with a vector of prices in which vector of prices each value represents an adjusted closing price for a business day ended in time-step $t_j$ ($j=1$ to $n$). For a financial instrument $S_i$ the vector of prices, i.e., a signal/time-series, is as follows:

$$S_i(P_{t_1}, P_{t_2}, P_{t_3}, P_{t_j}, ... P_{t_n})$$
$$...$$
$$S_m(P_{t_1}, P_{t_2}, P_{t_3}, P_{t_j}, ... P_{t_n})$$
  3.1

For all financial instruments, generate vectors representing the change in price for every two subsequent trading days:

$$S_i(\frac{P_{t_2}}{P_{t_1}}-1, \frac{P_{t_3}}{P_{t_2}}-1, \frac{P_{t_4}}{P_{t_3}}-1, ... \frac{P_{t_n}}{P_{t_{n-1}}}-1)$$
$$...$$
$$S_m(\frac{P_{t_2}}{P_{t_1}}-1, \frac{P_{t_3}}{P_{t_2}}-1, \frac{P_{t_4}}{P_{t_3}}-1, ... \frac{P_{t_n}}{P_{t_{n-1}}}-1)$$
  3.2

To simplify the representation of 3.2 it is presented as:

$$S_i(C_1, C_2, C_3, ... C_n)$$
$$...$$
$$S_m(C_1, C_2, C_3, ... C_n)$$
  3.3

The representation for the financial instruments $S_1, S_2, S_i, ... S_m$ as in expression 3.3 facilitates comparing between them because this representation is price-scale and value-scale independent. Since $n$ could be large (*e.g.*, if classification for one decade is desired), time slices of a constant size $h$ are defined. $h$ represents a set of $C$ values (see 3.3 expressions). Here, $h = 5$, representing five business days (one week). Presenting 3.3 expressions as a collection of time slices of length $h = 5$ results:

$$S_i[C_1,C_2,C_3,C_4,C_5]_1, S_i[C_6,C_7,C_8,C_9,C_{10}]_2, ... S_i[C_{n-4},C_{n-3},C_{n-2},C_{n-1},C_n]_k$$
$$...$$
$$S_m[C_1,C_2,C_3,C_4,C_5]_1, S_m[C_6,C_7,C_8,C_9,C_{10}]_2, ... S_m[C_{n-4},C_{n-3},C_{n-2},C_{n-1},C_n]_k$$

3.4

where the size of the total time-range of $n$ time-steps, also equals to $k$ time slices each of length of $h = 5$. The following representation, for example, is considered for the first time slice ($k = 1$):

$$S_i[C_1,C_2,C_3,C_4,C_5]_1$$
$$InverseS_i[-C_1,-C_2,-C_3,-C_4,-C_5]_1$$
$$...$$
$$S_m[C_1,C_2,C_3,C_4,C_5]_1$$
$$InverseS_m[-C_1,-C_2,-C_3,-C_4,-C_5]_1$$

3.5

In the classification problem considered here no labels are available for the signals and there is no information on how to refer to a set of values associated with a certain time slice. As such, a numerical value representing each signal is generated and assigned as the label of the signal. The numerical value labels denoted as $LS_i$ and $InverseLS_i$ are calculated for each signal:

$$LS_i = \sum_{l=1}^{h} S_i(C_l)$$

$$InverseLS_i = -\sum_{l=1}^{h} S_i(C_l)$$

... 3.6

$$LS_m = \sum_{l=1}^{h} S_m(C_l)$$

$$InverseLS_m = -\sum_{l=1}^{h} S_m(C_l)$$

The representation of self-labeling as shown in 3.6 expressions facilitates the application of supervised learning methods on unlabeled data sets. This is achieved by providing a supervised learning classification algorithm with pairs of adjusted representations of original signals (as shown as an example for $k=1$ in 3.5 expressions) and the adjusted representations' corresponding self-generated label (3.6 expressions).

## 3.2 Decision Tree Learning

The procedure described through expressions 3.1 - 3.6 is applied by acting several tables stored in a classifying database. Values according to expressions 3.5 and their corresponding labels as in expressions 3.6 are served as an input for a standard supervised learning algorithm. The supervised learning algorithm used is a decision tree algorithm. For each time slice, a decision tree is generated. A decision tree is a data structure that consists of branches and leaves. Leaves (also denoted as "nodes") represent classifications, and branches represent conjunctions of features that lead to those classifications. Each node has a unique title to distinguish the node from other nodes that the tree is composed. A node contains two or more records. Each record represents a financial instrument, its feature values (3.2 expressions) and its predictor value (3.6 expressions). The fewer financial instrument records in a node (the minimum is two), the less this node varies, i.e., a node with fewer records is more likely to represent a better classification between the financial instruments that the node contains.

    7,881 financial instruments are considered for classification including *NASDAQ*, *NYSE*, and *AMEX* and several market indices. To train the

algorithm with time-series and their inverses, expressions 3.5 - 3.6 are implemented. The implementation of expressions 3.5 - 3.6 doubles the size of the input instances to a total of 15,762 time-series. For the amount of data considered here, a typical size for one decision tree is in the range of approximately 2,000 to 8,000 nodes.

The decision tree classification results for any time slice considered, excluding noisy data, are stored in table of classification results of the classifying database. For the amount of data considered here, the number of records representing the nodes of one decision tree classification results is in the range of approximately 20,000 to 70,000 records. The procedure repeats itself with the next time slice until all of the 52 time slices of 2011 are processed and decision trees are created for them and added in a tabular format to the table of classification results. For the amount of data considered here, the number of records in the table is approximately 2.6 million.

### 3.3 Signal Composition

To receive classification results from classifying database, a similarity ranking algorithm is applied (figure 3). Consider a financial instrument specified by the user - the financial instrument is denoted as $S$ and the time-range is represented by a set of $t$ decision trees each representing one time slice classification. Note that, nodes with variability of predictors above a pre-defined threshold are filtered out as shown in figure 2.

---

Given a set of $T_1, T_2, \ldots T_t$ trees
   For each tree $T_i$ ($i = 1$ to $t$) each contains $N(T_i)$ nodes
      Find all $k$ nodes $N_j(T_i)$ ($j = 1$ to $k$) that contain $S$
         Find financial instruments in a node and increase by 1 a
         counter value associated with each financial instrument.
Sort the financial instruments in a descending order according to the total counter value of a financial instrument.

---

*Figure 3.* Similarity ranking algorithm.

## 4. Conclusion

The paper presents new machine learning methods: *signal composition*, which classifies time-series regardless of length, type, and quantity; and *self-labeling*, which is a supervised-learning enhancement. The methods were implemented on a financial use case as a search engine. The methods and the system were used to classify time-series of 7,881 American financial instruments traded at *NASDAQ*, *NYSE*, and *AMEX*, including several market indices (*e.g.*, *DJI*, *FTSE*, *HIS*) traded during 2011. The search engine allows a user to specify a particular financial instrument and receive in real-time a list of financial instruments that behave inversely to the particular financial instrument. The presented search approach of a cross-stock exchange classification assists the user to make diversification decisions in his or her portfolio.

The main objective to use time slices was to reduce the computation complexity - in practice, using too large number of input features in a classification algorithm may result unfeasible processing times. Splitting a signal into short time slices, performing classification for the shorter time slices separately and then applying a signal composition method, provides feasible processing times. Another reason to use time slices is the expected improved classification accuracy for certain problems.

Although the paper describes an implementation that relates to classification of financial instruments, the methods described could be implemented on other use cases. For example, the proposed methods and system could be applied to a series of non-financial behavioral patterns such as seismic or bio-medical patterns.

**References**

Haselsteiner, E. and Pfurtscheller, G. (2000). Using time-dependent neural networks for EEG classification. *IEEE Transactions on Rehabilitation Engineering*. 8(4), 457-463.

Hüsken, M. and Stagge, P. (2003). Recurrent neural networks for time-series classification. *Neurocomputing*. 50(C), 223-235.

Jović, A., Brkić, K., and Bogunović, N. (2012). Decision tree ensembles in biomedical time-series classification. *Pattern Recognition Lecture Notes in Computer Science*. 7476, 408-417.

Keogh, E. (2008). The UCR time-series classification/clustering homepage. Available: http://www.cs.ucr.edu/~eamonn/time_series_data/

Lines, J., Davis, L.M., Hills, J., and Bagnall, A. (2012). A shapelet transform for time-series classification. *Proceedings of the 18th ACM SIGKDD International Conference on Knowledge Discovery and Data Mining (KDD'12)* (pp. 289-297).

Nguyen, M.N. Li, X-L., and Ng, S-K. (2011). Positive unlabeled learning for time-series classification. *Proceedings of the Twenty-Second International Joint Conference on Artificial Intelligence (IJCAI)*. (pp. 1421-1426).

Perng, C-S., Wang, H., Zhang, S.R., and Parker, D.S. (2000). Landmarks: a new model for similarity-based pattern querying in time-series databases. *Proceedings of the 16th International Conference on Data Engineering*. (pp. 33-42).

Povinelli, R.J., Johnson, M.T., Lindgren, A.C., and Ye, J. (2004). Time-series classification using Gaussian mixture models of reconstructed phase spaces. *IEEE Transactions on Knowledge and Data Engineering*. 16(6), 779-783.

Ratanamahatana, C.A. and Wanichsan, D. (2008). Stopping criterion selection for efficient semi-supervised time-series classification. *Software Engineering, Artificial Intelligence, Networking and Parallel/Distributed Computing*. (pp. 1-14).

Wei, L. and Keogh, E. (2006). Semi-supervised time-series classification. *Proceedings of the 12th ACM SIGKDD International Conference on Knowledge Discovery and Data Mining (KDD'06)*. (pp. 748-753).

Wei, L. (2007). Self-training dataset. Available:
http://alumni.cs.ucr.edu/~wli/selfTraining/

Ye, L. and Keogh, E. (2009). Time-series shapelets: a new primitive for data mining. *Proceedings of the 15th ACM SIGKDD International Conference on Knowledge Discovery and Data Mining (KDD'09)*. (pp. 947-956).